\PassOptionsToPackage{hyphens}{url}

\documentclass[sigconf,authorversion]{acmart}
\makeatletter
\renewcommand\@formatdoi[1]{\ignorespaces}
\makeatother

\usepackage{fancyvrb}
\usepackage{listings}
\usepackage{colortbl}
\usepackage[relative,overlay]{textpos}
\usepackage{tikz}
\usepackage{hhline}
\usepackage{breakurl}

\hypersetup{
    breaklinks=true
}

\newcommand*\circled[1]{\tikz[baseline=(char.base)]{
    \node[shape=circle,fill=black, text=white,draw,inner sep=2pt] (char) {#1};}}

\lstdefinestyle{base}{
  language=Python,
  emptylines=1,
  breaklines=true,
  basicstyle=\ttfamily\color{black},
  moredelim=**[is][\color{red}]{@}{@},
  moredelim=**[is][\color{blue}]{|}{|}
}

\AtBeginDocument{%
  \providecommand\BibTeX{{%
    \normalfont B\kern-0.5em{\scshape i\kern-0.25em b}\kern-0.8em\TeX}}}

\setcopyright{acmcopyright}
\copyrightyear{2023}
\acmYear{2023}
\acmDOI{XXXXXXX.XXXXXXX}

\acmConference[ITiCSE '23]{28th annual ACM conference on Innovation and Technology in Computer Science Education}{July 10--12, 2023}{Turku, Finland}
\begin{document}

%%
%% The "title" command has an optional parameter,
%% allowing the author to define a "short title" to be used in page headers.
\title{Can Generative Pre-trained Transformers (GPT) Pass Assessments in Higher Education Programming Courses?}

%%
%% The "author" command and its associated commands are used to define
%% the authors and their affiliations.
%% Of note is the shared affiliation of the first two authors, and the
%% "authornote" and "authornotemark" commands
%% used to denote shared contribution to the research.
\author{Jaromir Savelka}
%\authornote{Both authors contributed equally to this research.}
\orcid{0000-0002-3674-5456}
\affiliation{%
  \institution{Carnegie Mellon University}
  \city{Pittsburgh}
  \state{PA}
  \country{USA}
}
\email{jsavelka@cs.cmu.edu}

\author{Arav Agarwal}
%\authornote{Both authors contributed equally to this research.}
\affiliation{%
  \institution{Carnegie Mellon University}
  \city{Pittsburgh}
  \state{PA}
  \country{USA}
}
\email{arava@andrew.cmu.edu}

\author{Christopher Bogart}
%\authornote{Both authors contributed equally to this research.}
\affiliation{%
  \institution{Carnegie Mellon University}
  \city{Pittsburgh}
  \state{PA}
  \country{USA}
}
\email{cbogart@andrew.cmu.edu}

\author{Yifan Song}
%\authornote{Both authors contributed equally to this research.}
\affiliation{%
  \institution{Carnegie Mellon University}
  \city{Pittsburgh}
  \state{PA}
  \country{USA}
}
\email{yifanson@andrew.cmu.edu}

\author{Majd Sakr}
%\authornote{Both authors contributed equally to this research.}
\affiliation{%
  \institution{Carnegie Mellon University}
  \city{Pittsburgh}
  \state{PA}
  \country{USA}
}
\email{msakr@cs.cmu.edu}

%%
%% By default, the full list of authors will be used in the page
%% headers. Often, this list is too long, and will overlap
%% other information printed in the page headers. This command allows
%% the author to define a more concise list
%% of authors' names for this purpose.
\renewcommand{\shortauthors}{Savelka and Agarwal, et al.}

%%
%% The abstract is a short summary of the work to be presented in the
%% article.
\begin{abstract}
  We evaluated the capability of generative pre-trained transformers~(GPT), to pass assessments in introductory and intermediate Python programming courses at the postsecondary level. Discussions of potential uses (e.g., exercise generation, code explanation) and misuses (e.g., cheating) of this emerging technology in programming education have intensified, but to date there has not been a rigorous analysis of the models' capabilities in the realistic context of a full-fledged programming course with diverse set of assessment instruments. We evaluated GPT on three Python courses that employ assessments ranging from simple multiple-choice questions~(no code involved) to complex programming projects with code bases distributed into multiple files (599 exercises overall). Further, we studied if and how successfully GPT models leverage feedback provided by an auto-grader. We found that the current models are not capable of  passing the full spectrum of assessments typically involved in a Python programming course (<70\% on even entry-level modules). Yet, it is clear that a straightforward application of these easily accessible models could enable a learner to obtain a non-trivial portion of the overall available score (>55\%) in introductory and intermediate courses alike. While the models exhibit remarkable capabilities, including correcting solutions based on auto-grader's feedback, some limitations exist (e.g., poor handling of exercises requiring complex chains of reasoning steps). These findings can be leveraged by instructors wishing to adapt their assessments so that GPT becomes a valuable assistant for a learner as opposed to an end-to-end solution.
  
  %\TODO{[MS] Instead of looking forward and disruption, might be better if you provide more details about where it does well and where it doesn’t do well.}%Given the rapid pace of improvement of generative pre-trained transformers, and their recent wide and well-publicized availability at minimal or no cost, we conclude that the technology is likely to disrupt the current established models of teaching programming at higher education institutions in the near future.
\end{abstract}

%%
%% The code below is generated by the tool at http://dl.acm.org/ccs.cfm.
%% Please copy and paste the code instead of the example below.
%%


%%
%% Keywords. The author(s) should pick words that accurately describe
%% the work being presented. Separate the keywords with commas.

% A "teaser" image appears between the author and affiliation
% information and the body of the document, and typically spans the
% page.
% \begin{teaserfigure}
%   \includegraphics[width=\textwidth]{iticse-teaser}
%   \caption{Seattle Mariners at Spring Training, 2010.}
%   \Description{Enjoying the baseball game from the third-base
%   seats. Ichiro Suzuki preparing to bat.}
%   \label{fig:teaser}
% \end{teaserfigure}

% \received{20 February 2007}
% \received[revised]{12 March 2009}
% \received[accepted]{5 June 2009}

%%
%% This command processes the author and affiliation and title
%% information and builds the first part of the formatted document.
\maketitle

\section{Introduction}
This paper explores the ability of generative pre-trained transformers (GPT), specifically \verb|text-davinci-003|, to pass typical assessments, such as multiple-choice question (MCQ) quizzes and coding tasks, routinely employed in introductory and intermediate programming courses. From three existing Python courses, we manually collected a sizeable data set of 599 exercises containing both, MCQs and coding tasks. We subjected GPT's answers and coding task solutions to the same assessment regimen as if they were coming from a human learner, including providing the model with an auto-grader's feedback and an opportunity to iterate on the solution. Based on the outcomes of the assessment we determine if the automatically generated work is of such quality that it would enable a human learner to successfully complete the courses.

The release of ChatGPT\footnote{OpenAI: ChatGPT. Available at: \url{https://chat.openai.com/} [Accessed 2023-01-20]} has made the general public aware of the capabilities of GPT models to write fluent text and answer questions. Although, the underlying GPT-3 technology has been available since 2020 \cite{brown2020language}, the ready availability of this more easy-to-use version of the tool is spurring educators to anticipate how educational methods will be forced to change in the near future. There are already early examples of measures educational institutions are taking. New York City public schools have recently blocked ChatGPT from the school networks and school-owned devices~\cite{ElsenRooney2023}, out of concern not only that students would use the tool to complete assignments, but that it sometimes produces unreliable or inappropriate content. University classes are revising methods to include more in-class writing~\cite{Huang2023}. They are also exploring tools for detecting AI-generated text, such as GPTZero~\cite{Bowman2023}.

Many educators in CS, and programming instructors in particular, have been aware of this development for some time. One of GPT's strongest capabilities is computer program synthesis. A number of code generation tools such as OpenAI's Codex~\cite{https://doi.org/10.48550/arxiv.2107.03374}, DeepMind's AlphaCode~\cite{doi:10.1126/science.abq1158}, or Amazon's CodeWhisperer~\cite{ankur2022} have been released recently. GitHub's Copilot\footnote{GitHub Copilot: Your AI pair programmer. Available at: \url{https://github.com/features/copilot} [Accessed 2023-01-20]} (a version of Codex) likely attracted the most attention due to its seamless integration with mainstream IDEs, such as Visual Studio Code. It is 
%marketed as ``Your AI pair programmer'' (a reference to pair programming \cite{beck2000extreme,mcdowell2002effects}) and 
available for free to students and educators. While there are technological or cost barriers to use the other tools, these are non-existent in case of Copilot or ChatGPT. Hence, it is inevitable that learners will tend to utilize such tools when completing their course assignments.

% In this paper we attempt to comprehensively evaluate how well a state-of-the-art LLM can perform on all aspects of an introductory course, from simple multiple-choice formative reading comprehension questions to in-depth programming projects expected to take students multiple days to complete.  Empirical data about an LLM's performance should inform further research by ruling out both over-estimates of LLM's capabilities (i.e. that they can effortlessly and perfectly complete all assignments), or under-estimates (i.e. that the known flaws in their capabilities will make them unusable by students).

% Finally, we suggest ways educational designers can, on one hand, LLM-proof their assignments, and on the other hand, find positive ways to employ LLM, such as testing their own course material for LLM-vulnerability, generating ideas and lesson plans, and encouraging critical thinking in students by helping them engage with LLMs responsibly.

% \subsection{MCQ Answering and Coding Tasks}

To investigate if and how GPT could pass a battery of diverse assessments to successfully complete a full-fledged Python programming course, we analyzed the following research questions:

\begin{itemize}
    \item[(RQ1)] How reliably can GPT generate correct answers to MCQ assessments employed in such courses?
%    \item[(RQ2)] What are some common properties of MCQs that are answered (in)correctly by GPT? 
    \item[(RQ2)] How reliably can GPT produce solutions to complex Python coding tasks from instructions designed for humans?
%    \item[(RQ4)] What are some common properties of coding tasks and/or instructions that are handled (un)successfully by GPT?
    \item[(RQ3)] How well can GPT utilize feedback generated by an auto-grader in order to improve defective solutions?
\end{itemize}

By carrying out this work, we provide the following contributions to the CS education research community. To our best knowledge, this is the first comprehensive study that:

\begin{itemize}
    \item[(C1)] evaluates performance of GPT models on the \emph{full spectrum of diverse assessment instruments} employed by real existing Python programming courses.
    \item[(C2)] evaluates performance of GPT models on \emph{MCQ-style assessments that involve code snippets}.
    \item[(C3)] evaluates the capability of GPT models to correct a solution of a coding task in \emph{response to an auto-grader's feedback}.
\end{itemize}

% \begin{itemize}
%     \item Evidence that the current state-of-the-art GPT models are not yet capable of passing full spectrum of assessments in the quality sufficient to successfully complete a typical Python programming course.
%     \item Evidence that a GPT model can obtain a considerable portion of the available score by passing diverse types of assessments.
%     \item Evidence that 
%     \item Reproducible workflow to assess 
% \end{itemize}

\section{Related Work}
\label{sec:related_work}
%In the literature surrounding code generation via large language models, there are several works that evaluate LLMs' effectiveness at solving different kinds of evaluations intended for humans.
There are works evaluating the performance of GPT models on MCQ data sets from various domains. Robinson et al. \cite{robinson2022} apply InstructGPT \cite{ouyang2022training} and Codex to OpenBookQA~\cite{mihaylov2018can}, StoryCloze \cite{mostafazadeh2016corpus}, and RACE-m \cite{lai2017race} data sets which focus on multi-hop reasoning, recall, and reading comprehension, reporting 77.4-89.2\% accuracy. Hendryks et al. \cite{hendrycks2022} created data set that includes a wide variety of MCQs across STEM, humanities, and arts, with GPT performing at 43.9\% accuracy. Lu et al. \cite{pan2022} collected a data set of 21,208 MCQs reporting the accuracy of 74.04\%. Zong and Krishnamachari applied GPT to a data set of math word problems MCQs, achieving 31\%  accuracy \cite{zong2022solving}. Drori and Verma \cite{https://doi.org/10.48550/arxiv.2111.08171} used Codex to write Python programs to solve 60 computational linear algebra MCQs (100\% accuracy). Other applications of GPT to solve MCQ-based exams include United States Medical Licensing Examination (USMLE)~\cite{kung2022performance, Gilson2022HowWD, Lievin2022CanLL}, Multistate Bar Examination \cite{bommarito2022gpt}, and American Institute of Certified Public Accountants' (AICPA) Regulation exam \cite{bommarito2023gpt}. Jalil et al. studied the performance of ChatGPT on open-ended questions on an advanced programming topic \cite{jalil2023chatgpt}.

There is an emerging body of work that explores the effectiveness of code generation tools when applied to programming assignments in higher education courses. Finnie-Ansley et al. evaluated Codex on 23 programming tasks used as summative assessments in a CS1 programming course reporting 43.5\% success rate; 72\% on re-attempts \cite{10.1145/3511861.3511863}. Finnie-Ansley et al. also studied Codex' performance on CS2 tasks \cite{finnie2023my}. Wermelinger~\cite{wermelinger2023using} states that Copilot can be a useful springboard towards solving CS1 problems while learners' substantial contribution is still required. Denny et al.~\cite{https://doi.org/10.48550/arxiv.2210.15157} focused on the effects of prompt engineering when applying Copilot to a set of 166 exercises from the publicly available CodeCheck repository.\footnote{CodeCheck: Python Exercises. Available at: \url{https://horstmann.com/codecheck/python-questions.html} [Accessed 2022-01-22]} They observed a sizeable improvement of the 47.9\% success rate on the initial submission to 79.0\% when instructions rewording was used. Biderman and Raff \cite{Biderman2022FoolingMD} demonstrated how GPT solves introductory coding tasks with code alterations bypassing MOSS detection on Karnalim et al's \cite{karnalim2019source} plagiarism data set. Becker et al. \cite{becker2022programming} discuss the opportunities and challenges posed by the code generating tools.

Outside of the educational context, there have been studies exploring GPT's capabilities on competitive and interview programming tasks. Chen et al. \cite{https://doi.org/10.48550/arxiv.2107.03374} released the HumanEval data set, consisting of 164 hand-written problems, where Codex achieved 28.8\% success rate on the first attempt and 72.3\% when allowed 100 attempts. Li et al. \cite{doi:10.1126/science.abq1158} report Deepmind's AlphaCode performance on Codeforces competitions,\footnote{Codeforces. Available at: \url{https://codeforces.com/contests} [Accessed 2023-01-22]} achieving a 54.3\% ranking amongst 5,000 participants. Karmakar et al. \cite{Karmakar2022CodexHH} reported 96\% pass rate for Codex on a data set of 115 programming problems from HackerRank.\footnote{HackerRank. Available at: \url{https://www.hackerrank.com/} [Accessed 2023-01-22]} Nguyen and Nadi \cite{9796235} reported Copilot's effectiveness on LeetCode\footnote{LeetCode. Available at: \url{https://leetcode.com/} [Accessed 2023-01-22]} problems, achieving 42\% accuracy with no prompt tuning. Some researchers focus on analyzing specific weaknesses of the code synthesis tools (presence of common bad patterns \cite{10006873}, common security problems \cite{pearce2022asleep}, or  code that is challenging for humans to debug \cite{10.1145/3491101.3519665, Barke2022GroundedCH}).

% Need to verify
%Our work is unique in three regards. Firstly, it uses a mixed and complete dataset of multiple-choice questions and programming questions, asking Codex to solve every exercise in a course. Secondly, unlike other works which generate multiple potential solutions from Codex to then evaluate on existing work, we give Codex the auto-grader feedback from the last solution, and ask it to generate a new solution from it. The goal of this is to more closely simulate how a student is likely to use the software. Lastly, we ask Codex to generate multiple solutions together, asking what is the minimal amount of effort necessary to generate a correct solution on the student’s behalf.

\section{Data Set}
We manually collected assessment exercises from three Python programming courses. \emph{Python Essentials - Part 1 (Basics)}\footnote{OpenEDG: Python Essentials - Part 1 (Basics). Available at: \url{https://edube.org/study/pe1} [Accessed 2023-01-15]} (\textbf{PE1}) aims to guide a learner from a state of complete programming illiteracy to a level of programming knowledge which allows them to design, write, debug, and run programs encoded in the Python language. \emph{Python Essentials - Part 2 (Intermediate)} (\textbf{PE2})\footnote{OpenEDG: Python Essentials - Part 2 (Intermediate). Available at: \url{https://edube.org/study/pe2} [Accessed 2023-01-15]} is focused on more advanced aspects of Python programming, including modules, packages, exceptions, file processing, object-oriented programming. Finally, \emph{Practical Programming with Python}\footnote{Sail(): Social and Interactive Learning Platform. Available at: \url{https://sailplatform.org/courses} [Accessed 2023-03-09]} (\textbf{PPP}) emphasizes hands-on experience with fundamental Python constituents and exposure to software development tools, practices, and real-world applications.
%\footnote{Sail(): Courses. Available at: \url{https://sailplatform.org/courses} [Accessed 2023-01-15]}
%While PE1 and PE2 are available to the general public at no cost, PPP is being offered at various types of higher education institutions across the United States.

%The tests determine if learners pass the courses whereas quizzes are meant as practice. 

\begin{table}
  \caption{Descriptive statistics of the created dataset. Each row provides information about the assessments each of the courses employ. Each column reports on the distribution of the assessment type accross the courses.}
  \label{tab:dataset}
  \setlength{\tabcolsep}{5pt}
  %\footnotesize
  \begin{tabular}{l|r|rrrr}
  \toprule
    Course        & Units    & MCQ     & MCQ     & Coding    & Course\\
                  & (topics) & (plain) & (+code) & Exercise  & Overall \\
  \hline
    PE1           & 4        & 53          & 96          & -          &\bf 149  \\
    PE2           & 4        & 65          & 83          & -          &\bf 148  \\
    PPP           & 8        & 89          & 144         & 69         &\bf 302  \\
  \hline
    Type Overall  & 16       & \bf 207     & \bf 323     &\bf 69      &\bf 599  \\
  \bottomrule
  \end{tabular}
\end{table}

PE1 and PE2 employ MCQ-style assessments. Formative assessments are called quizzes while summative assessments are called tests. The MCQs often include small snippets of code for learners to reason about. From the two courses, we collected 297 questions (179 have code snippets). Table~\ref{tab:dataset} has additional details.
PPP uses MCQ-style inline activities as formative assessment and tests as summative assessment. PPP mostly employs the project/problem-based education model \cite{kokotsaki2016project} where learners individually work on large programming projects subdivided into tasks that require sustained effort extending over several days. Projects are auto-graded providing learners immediate actionable feedback (examples in Figures \ref{fig:example-success} and \ref{fig:example-fail} of Section \ref{sec:discussion}). Learners can iterate on their solutions until a project deadline. The score from the projects, tests, and reflections (discussion posts) determines if a learner successfully completes the course. We collected 233 MCQs (144 with code snippets) and 69 coding activities (elements of the 32 project tasks). Further details are reported in Table~\ref{tab:dataset}.

% \begin{figure}
% \footnotesize
% \begin{verbatim}
% [RULE] There should be the `retrieve_popular_books` function in the
% `books_retrieval.py` file that upon invocation generates a JSON file 
% listing popular and highly rated books. 
% [RESULT] FAILED (0/15)
% [FEEDBACK] The value under the `33c05332-c53a-495d-b95d-f6a7f5400b4a` 
% book id's `info` and `num_pages` keys is of type `<class 'str'>` but it 
% should be of type `<class 'int'>`.
% \end{verbatim}
% \caption{Sail() Project Task Feedback Snippet}
% \label{fig:sail_feedback}
% \end{figure}

\section{Model}
\label{sec:models}
We used \verb|text-davinci-003|, one of the most advanced GPT models offered by OpenAI. Since their introduction in 2017, transformer architectures \cite{vaswani2017attention} have revolutionized numerous areas of machine learning (ML) research and have greatly impacted many applications of natural language processing (NLP) and computer vision~(CV). The \verb|text-davinci-003| model builds on top of previous InstructGPT model (\verb|text-davinci-002|), which in turn is based on \verb|code-davinci-002| (focused on code-completion tasks).\footnote{OpenAI: Model index for researchers. Available at: \url{https://beta.openai.com/docs/model-index-for-researchers/instructgpt-models} [Accessed 2023-01-15]} %Recently, OpenAI has also released ChatGPT which reportedly resulted in over 1M user sign-ups in just 5 days. 

For MCQs, we conventionally (Section \ref{sec:related_work}) set the \verb|temperature| to 0.0 (no randomness in the output), \verb|max_tokens| to 500 (token roughly corresponds to a word), \verb|top_p| to 1, \verb|frequency_penalty| to 0, and \verb|presence_penalty| to 0. Using the value of 0.0 for temperature is in line with existing work \cite{Lievin2022CanLL, bommarito2022gpt}.
%, and also the general observation that, as we are primarily looking at singular outputs for the models, the most likely output at higher temperatures is going to be the greedy output. 
For coding tasks, we set the max\_tokens to 2,000 (solutions could be extensive) and temperature to 0.7. The use of \verb|text-davinci-003| for code generation is somewhat untypical as is setting the temperature to 0.7. Prior literature mostly uses Codex (\verb|code-davinci-002|) and it is recommended to set the temperature to 0.0. We compared \verb|text-davinci-003| against Codex on a large number of coding generation problems and observed similar performance. Hence, we opted for \verb|text-davinci-003| in order to use the same model for both, MCQs and coding tasks. As for the temperature, the default 0.7 is used to encourage the model to make changes based on the provided feedback. Our investigation also suggests that \verb|text-davinci-003| is quite robust in terms of its ability to pass the coding task assessments irrespective of the temperature setting. The older \verb|code-davinci-002| and \verb|text-davinci-002| models appear to be somewhat more sensitive.

\section{Experimental Design}
To test the performance on MCQs, we submit questions one by one using the \verb|openai| Python library\footnote{GitHub: OpenAI Python Library. Available at: \url{https://github.com/openai/openai-python} [Accessed 2023-01-16]} which is a wrapper for the OpenAI's REST API. We embed each question in the prompt template shown in Figure \ref{fig:mcq-prompt-template}. The model returns one or more of the choices as the prompt completion, which is then compared to the reference answer. Following the approach adopted by PE1 and PE2, partially correct answers are considered to be incorrect.

\begin{figure}
\footnotesize
\begin{Verbatim}[frame=single,commandchars=\\\{\}]
I am a highly intelligent bot that can easily handle answering
multiple-choice questions on introductory Python topics. Given a
question and choices I can always pick the right ones.

Question: \textcolor{blue}{\string{\string{question\string}\string}}

Choices:
\textcolor{blue}{\string{\string{choices\string}\string}}

The correct answer:
\end{Verbatim}
\begin{textblock*}{3.4cm}(5.7cm,-2.4cm)
\circled{1}
\end{textblock*}
\begin{textblock*}{3.4cm}(1.5cm,-1.95cm)
\circled{2}
\end{textblock*}
\begin{textblock*}{3.4cm}(0.1cm,-1.1cm)
\circled{3}
\end{textblock*}
\caption{MCQ Prompt Template. The text of the preamble (1) is inspired by OpenAI's QA example. The \string{\string{question\string}\string} token~(2) is replaced with the question text. The \string{\string{choices\string}\string} token~(3) is replaced with the candidate answers where each one is placed on a single line preceded by a capital letter.}
\label{fig:mcq-prompt-template}
\end{figure}

\begin{figure}
\footnotesize
\begin{Verbatim}[frame=single,commandchars=\\\{\}]
TASK
Implement a Python program to print "Hello, World!" in hello.py.
=== hello.py ===
# TODO 1
===

SOLUTION
=== hello.py ===
print("Hello, World!")
===

TASK
\textcolor{blue}{\string{\string{instructions\string}\string}}
=== \textcolor{blue}{\string{\string{file_name\string}\string}} ===
\textcolor{blue}{\string{\string{handout\string}\string}}
===

SOLUTION
\end{Verbatim}

\begin{textblock*}{3.4cm}(4.8cm,-3.7cm)
\begin{Verbatim}[frame=single,commandchars=\\\{\}]
[...]

SOLUTION
=== \textcolor{blue}{\string{\string{file_name\string}\string}} ===
\textcolor{blue}{\string{\string{solution\string}\string}}
===

FEEDBACK
\textcolor{blue}{\string{\string{feedback\string}\string}}

FIXED SOLUTION
\end{Verbatim}
\end{textblock*}
\begin{textblock*}{1.4cm}(1.5cm,-4.05cm)
\circled{1}
\end{textblock*}
\begin{textblock*}{1.4cm}(1.7cm,-1.95cm)
\circled{2}
\end{textblock*}
\begin{textblock*}{1.4cm}(2.3cm,-1.66cm)
\circled{3}
\end{textblock*}
\begin{textblock*}{1.4cm}(1.06cm,-1.33cm)
\circled{4}
\end{textblock*}
\begin{textblock*}{1.4cm}(7.1cm,-2.75cm)
\circled{5}
\end{textblock*}
\begin{textblock*}{1.4cm}(5.95cm,-2.45cm)
\circled{6}
\end{textblock*}
\begin{textblock*}{1.4cm}(6.0cm,-1.33cm)
\circled{7}
\end{textblock*}

\caption{Coding Task Prompt Templates. Outer frame is the first submission template. The preamble (1) primes the model to generate a solution code as completion. The \string{\string{instructions\string}\string} token~(2) is replaced with the coding task instructions. A starter code is injected into \string{\string{file\_name\string}\string} (3) and \string{\string{handout\string}\string} (4) tokens. The inner frame shows the template for re-submission (appended to the original). The \string{\string{file\_name\string}\string}~(5) and \string{\string{solution\string}\string} (6) were replaced with the GPT's solution and the \string{\string{feedback\string}\string} (7) with the auto-grader's feedback.}
\label{fig:project-prompt-template}
\end{figure}

% When learners submit a quiz or a test they may review their answers and re-take the assessment. During the review they are provided with the information as to which questions were answered correctly and which were not. However, they are not provided with the correct answers. To simulate this procedure, we submit a follow-up request to the API, as a re-take, in case a question has been answered incorrectly. For the re-take, the question is embedded in the prompt template shown in Figure \ref{fig:mcq-prompt-retake-template}. For this prompt a different preamble is used as well as the originally generated incorrect answer.

% \begin{figure}
% \footnotesize
% \begin{verbatim}
% I am a highly intelligent bot that can easily correct a wrong answer to
% a multiple-choice question on introductory Python topics. Given a 
% question, a list of choices, and an incorrect answer I can always pick 
% the right ones.

% Question: {{question}}

% Choices:
% {{choices}}

% The incorrect answer: {{answer}}
% The correct answer:
% \end{verbatim}
% \caption{MCQ Re-take Prompt Template}
% \label{fig:mcq-prompt-retake-template}
% \end{figure}

In coding tasks, we submit the instructions using the prompt template shown in Figure \ref{fig:project-prompt-template}. Here, we encountered the model's limitation on the prompt length (4,097 tokens). 
%Within this limit, it was necessary to fit: (i) the prompt boilerplate; (ii) the instructions; (iii) the contents of the handout files (usually starter code) distributed to learners; and (iv) the solution generated by the model (i.e., the prompt completion). 
Instead of full project~(8) instructions we submitted the individual project tasks (32). If a task could not be fitted into a prompt, we decreased the \verb|max_tokens| parameter (space for solution) to <2,000. If this did not resolve the issue we edited the instructions leaving out pieces that could be reasonably expected as not being useful for GPT. As the last resort, we would split the task into several smaller coding activities (69 overall) if the task was to develop several loosely coupled elements. GPT's solution was  submitted to the auto-grader.

To each submission, the auto-grader assigns a score and generates detailed actionable feedback (examples shown in Figures \ref{fig:example-success} and \ref{fig:example-fail} of Section \ref{sec:discussion}). If the full score was not achieved we amended the original prompt with the addendum (template shown in Figure~\ref{fig:project-prompt-template}). Then, we submitted the revised solution to the auto-grader and repeated the process until either the full score was achieved or the solution remained unchanged from the preceding one (impasse).

\section{Results}
The results of applying GPT to the MCQ exercises are reported in Tables \ref{tab:pe1_results}, \ref{tab:pe2_results}, and \ref{tab:ppp_concepts_results}. To gauge the rate of improvement, we also report performance of the \verb|text-davinci-002| (preceding generation). The \verb|text-davinci-003| correctly answered 341 MCQs out of the 530 (64.3\%). The \verb|text-davinci-002| generated 301 correct answers~(56.8\%). Hence, we observe a sizeable improvement in the more recent model as compared to the older one. The results from PE1 are reported in Table \ref{tab:pe1_results}. %The course employs the 70\% passing threshold for all its assessments. 
In order to pass the course the score of 70\% or better is required from all the 5 tests. The more successful \verb|text-davinci-003| model passed only the first course module. It fell short on the remaining three and collected only 48.6\% from the Summary Test. This amounts to failing the course with the average score of 56.1\%. The performance of the models on PE2 is presented in Table \ref{tab:pe2_results}. The assessment scheme of PE2 is the same as that of PE1. Here, \verb|text-davinci-003| successfully passed 3/4 module tests. However, it also failed the Summary Test (65.0\%). This also amounts to failing the course with the average score of 67.9\%.

\begin{table}
  \caption{PE1 results. td-002 is text-davinci-002, td-003 is text-davinci-003, green indicates passing, and red  failure.}
  \label{tab:pe1_results}
  \footnotesize
  \begin{tabular}{lrrrr}
    \toprule
                 &\multicolumn{2}{c}{Quizzes} &\multicolumn{2}{c}{Tests} \\
    Module Topic                         & td-002     & td-003     & td-002    & td-003 \\
    \hline
    Introduction to Python and           &8/10         &10/10        &\cellcolor{red!15}5/10        &\cellcolor{green!15}9/10\\
    computer programming                 &(80.0\%)     &(100\%       &\cellcolor{red!15}(50.0\%)    &(\cellcolor{green!15}90.0\%)\\
    \hline
    Data types, variables, basic I/O,    &9/10         &10/10        &\cellcolor{red!15}10/20       &\cellcolor{red!15}10/20\\
    operations, and basic operators      &(90.0\%)     &(100\%)      &\cellcolor{red!15}(50.0\%)    &\cellcolor{red!15}(50.0\%)\\
    \hline
    Boolean values, conditionals, loops, &8/10         &7/10         &\cellcolor{red!15}11/20       &\cellcolor{red!15}12/20\\
    lists, logical and bitwise operators &(80.0\%)     &(70.0\%)     &\cellcolor{red!15}(55.0\%)    &\cellcolor{red!15}(60.0\%)\\
    \hline
    Functions, tuples, dictionaries,     &7/12         &9/12         &\cellcolor{red!15}14/22       &\cellcolor{red!15}12/22\\
    data processing and exceptions       &(58.3\%)     &(75.0\%)     &\cellcolor{red!15}(63.6\%)    &\cellcolor{red!15}(54.5\%)\\
    \hline
    Completion (Summary Test)            &-            &-            &\cellcolor{red!15}15/35       &\cellcolor{red!15}17/35\\
                                         &             &             &\cellcolor{red!15}(42.9\%)    &\cellcolor{red!15}(48.6\%) \\
    \hline
    \bf Course Total                     &\bf 32/42    &\bf 36/42    &\bf 55/107   &\bf 60/107\\
                                         &\bf (76.2\%) &\bf (85.7\%) &\bf (51.4\%) &\bf (56.1\%)\\
    \bottomrule
  \end{tabular}
\end{table}

\begin{table}
  \caption{PE2 results. td-002 is text-davinci-002, GPT-3.5 is text-davinci-003, green indicates passing, and red  failure.}
  \label{tab:pe2_results}
  \footnotesize
  \begin{tabular}{lrrrr}
    \toprule
                 &\multicolumn{2}{c}{Quizzes} &\multicolumn{2}{c}{Tests} \\
    Module Topic                         & td-002     & td-003     & td-002     & td-003 \\
    \hline
    Modules, packages, and PIP           &7/10         &6/10         &\cellcolor{red!15}12/18        &\cellcolor{green!15}14/18\\
                                         &(70.0\%)     &(60.0\%)     &\cellcolor{red!15}(66.7\%)     &\cellcolor{green!15}(77.8\%)\\
    \hline
    Strings, string and list methods,    &8/10         &6/10         &\cellcolor{green!15}11/15        &\cellcolor{green!15}11/15\\
    and exceptions                       &(70.0\%)     &(60.0\%)     &\cellcolor{green!15}(73.3\%)     &\cellcolor{green!15}(73.3\%)\\
    \hline
    Object-oriented programming          &7/10         &8/10         &\cellcolor{red!15}9/17         &\cellcolor{green!15}12/17\\
                                         &(70.0\%)     &(80.0\%)     &\cellcolor{red!15}(52.9\%)     &\cellcolor{green!15}(70.6\%)\\
    \hline
    Miscellaneous                        &9/12         &9/12         &\cellcolor{red!15}9/16         &\cellcolor{red!15}9/16 \\
                                         &(75.0\%)     &(75.0\%)     &\cellcolor{red!15}(56.2\%)     &\cellcolor{red!15}(56.2\%)\\
    \hline
    Completion (Summary Test)            &-            &-            &\cellcolor{red!15}26/40        &\cellcolor{red!15}26/40\\
                                         &             &             &\cellcolor{red!15}(65.0\%)     &\cellcolor{red!15}(65.0\%) \\
    \hline
    \bf Course Total                     &\bf 31/42    &\bf 29/42    &\bf 67/106   &\bf 72/106\\
                                         &\bf (73.8\%) &\bf (69.0\%) &\bf (63.2\%) &\bf (67.9\%)\\
    \bottomrule
  \end{tabular}
\end{table}

\begin{table}
  \caption{PPP MCQ results. No color coding in this table; see Table \ref{tab:ppp_overall_results} for info about passing the course modules.}
  \label{tab:ppp_concepts_results}
  \footnotesize
  \begin{tabular}{lrrrr}
    \toprule
                 &\multicolumn{2}{c}{Inline Activities} &\multicolumn{2}{c}{Tests} \\
    Module Topic                         & td-002     & td-003     & td-002    & td-003 \\
    \hline
    Python basics and introduction to    &20/30        &21/30        &7/12        &9/12 \\
    functions                            &(66.7.0\%)   &(70.0\%)     &(58.3\%)    &(75.0\%)\\
    \hline
    Control flow, strings, input and     &9/22         &10/22        &6/11        &8/11  \\
    output                               &(40.9\%)     &(45.5\%)     &(54.5\%)    &(72.7\%)\\
    \hline
    Python data structures               &8/18         &10/18        &6/14        &9/14 \\
                                         &(44.4\%)     &(55.6\%)     &(42.9\%)    &(64.3\%) \\
    \hline
    Object-oriented programming          &6/14         &7/14         &5/11        &10/11\\
                                         &(42.9\%)     &(50.0\%)     &(45.5\%)    &(90.9\%)\\
    \hline
    Software development                 &11/19        &12/19        &6/10        &7/10 \\
                                         &(57.9\%)     &(63.2\%)     &(60.0\%)    &(70.0\%) \\
    \hline
    Data manipulation                    &7/17         &9/17         &5/13       &5/13 \\
                                         &(41.2\%)     &(52.9\%)     &(38.5\%)   &(35.5\%) \\
    \hline
    Web scraping and office document     &2/10         &5/10         &3/5        &3/5 \\
    processing                           &(20.0\%)     &(50.0\%)     &(60.0\%)   &(60.0\%) \\
    \hline
    Data analysis                        &14/22        &17/22        &1/5         &2/5 \\
                                         &(63.6\%)     &(77.3\%)     &(20.0\%)    &(40.0\%) \\
    \hline
    \bf Course Total                     &\bf 77/152   &\bf 91/152   &\bf 39/81    &\bf 53/81\\
                                         &\bf (50.7\%) &\bf (59.9\%) &\bf (48.1\%) &\bf (65.4\%)\\
    \bottomrule
  \end{tabular}
\end{table}

\begin{table*}
  \caption{Coding tasks results. Prompt Tokens and Compl. Tokens columns report the length of the instructions  and task solutions. Max score is the maximum score achievable from a task. First score is the score after first submission. Attempts is the number of submissions before the full score or no-change impasse were reached. Final score is the score after feedback. The entries with + describe situations where a task was divided into multiple subactivities due to the prompt's length limitation.}
  \label{tab:project_results}
  \footnotesize
  \setlength{\tabcolsep}{3.5pt}
  \begin{tabular}{llrrrrrr}
    \toprule
    Project Topic &Tasks (skills) &Prompt Tokens & Compl. Tokens &Max Score &1st Score &Attempts &Final Score\\
    \midrule
    Types, variables, functions &Variable assignment                               &491             &226             &13 &13 &1 &13\\
                                &User-defined functions, imports, return vs print  &909             &302             &20 &16 &2 &16\\
                                &Simple scripts, comments                          &642+814+1,008   &364+543+570     &12+19+19 &12+19+19 &1+1+1 &12+19+19\\
                                &Testing                                           &1,307           &740             &12 &12 &1 &12\\
                                &\bf Project 1 Total                               &\bf 5,171       &\bf 2,745       &\bf 95 &\bf 91 (95.8\%) &\bf 6 &\bf 91 (95.8\%)\\
    \midrule
    Iteration, conditionals,    &Conditional statements                            &677             &261             &10        &10              &1     &10      \\
    \ strings, basic I/O        &Strings, while loop, for loop, complex printing   &771+1,048+1,405 &334+602+656     &11+12+12 &5+0+0 &3+2+2 &9+0+0\\
                                &Read and write files                              &795             &196             &15       &0     &3     &3\\
                                &Complex script                                    &2,710           &729             &35       &25    &2     &25\\
                                &\bf Project 2 Total                               &\bf 7,406       &\bf 2,778       &\bf 95 &\bf 40 (42.1\%) &\bf 13 &\bf 47 (49.5\%)\\
    \midrule
    Lists, sets, tuples and     &Create container, access, add, remove and update  &276+559+779+    &81+95+241+      &4+4+8+    &4+4+8+   &1+1+1+ &4+4+8+\\
    \ dictionaries              &\ elements, and convert containers across types   &+631+900+1,787  &+212+305+563    &+8+8+8    &+8+6+8   &+1+2+1 &+8+8+8\\
                                &Nested data structure, transformation, sorting,   &874+536+1,141+  &462+389+544+    &15+10+10+ &0+10+10+ &3+1+1  &0+10+10+\\
                                &\ export to file, complex report                  &+985+1,379      &+680+586        &+5+15     &+5+5     &+1+2   &+5+15\\
                                &\bf Project 3 Total                               &\bf 9,847       &\bf 4,158       &\bf 95    &\bf 68 (71.6\%) &\bf15 &\bf 80 (84.2\%)\\
    \midrule
    Classes, objects,           &Implement classes                                 &1,124           &445             &13     &13   &1     &13      \\
    \ attributes and methods    &Define, access and set private attributes         &1,581           &737             &19     &19   &1     &19      \\
                                &Inheritance                                       &2,033           &828             &20     &20   &1     &20      \\
                                &Implement re-usable utility functions             &722+1,489       &173+953         &4+12   &4+12 &1+1   &4+12     \\
                                &Composition, object instantiation                 &2,053+1,536     &1,241+115       &9+7    &9+7  &1+1   &9+7     \\
                                &Override special methods (repr, eq)               &1,861           &1,587           &11     &0    &3     &0       \\
                                &\bf Project 4 Total                               &\bf 12,399      &\bf 6,079       &\bf 95 &\bf 84 (88.4\%) &\bf 10 &\bf 84 (88.4\%)\\
    \midrule
    Debugging, refactoring,     &Identify and fix errors in code                   &1,979+2,030     &193+920         &5+5 &0+0 &2+2 &0+0 \\
    \ testing and packaging     &Refactor larger code base                         &-               &-               &14  &-   &-   &-   \\
                                &Exception handling                                &1,681           &592             &13  &4   &3   &4   \\
                                &Analyze and fix code on style correctness         &1,010+1,756+1,323+&324+667+450+    &6+6+6+ &0+0+0+ &3+3+3+ &0+0+0+\\
                                &                                                  &+2,124+3013        &+813+1050        &+6+4   &+0+0   &+3+3   &+0+0\\
                                &Test-driven development                           &989+547+1,630+  &452+305+622+    &5+5+5+ &0+0+0+ &3+2+2+ &0+5+0+\\
                                &                                                  &+1,939          &+563            &+5     &+0     &+2     &+0\\
                                &Package Python application using pip              &-               &-               &10  &-   &-   &-   \\
                                &\bf Project 5 Total                               &\bf 22,536         &\bf 8,516         &\bf 95 &\bf 4 (4.2\%) &\bf 31 &\bf 9 (9.5\%)\\
    \midrule
    Files and datastores        &Load and store data in files                      &934+903+850     &386+450+458     &15+15+15 &0+15+15  &4+1+1 &15+15+15\\
                                &Create SQL objects, load and query data in SQL    &689+384+761     &332+221+394     &10+10+10 &10+10+10 &1+1+1 &10+10+10\\
                                &Load and query data in MongoDB                    &1,067           &485             &20       &20       &1     &20\\
                                &\bf Project 6 Total                               &\bf 5,588       &\bf 2,726       &\bf 95 &\bf 80 (84.2\%) &\bf 10 &\bf 95 (100.0\%)\\
    \midrule
    Web scraping and office     &Get HTML, extract information from HTML,         &1,789+1,533      &470+1,030       &25+10 &0+5  &5+7  &10+8\\
    \ document processing       &\ handle multiple HTML files    \\
                                &Manipulate Excel files programatically            &1,676+1,559+1,688 &1,313+1,282+1,339 &10+5+10  &0+5+0  &3+1+3  &0+5+0\\
                                &Authenticate and utilize public API               &886             &387             &15       &0      &2      &0\\
                                &Manipulate Word files programmatically            &1,175           &717             &20       &10     &4      &20\\
                                &\bf Project 7 Total                               &\bf 10,306      &\bf 6,538       &\bf 95 &\bf 20 (21.1\%) &\bf 25 &\bf 43 (45.3\%)\\
    \midrule
    Data analysis               &Load data to pandas, merge pandas DataFrames,         &2,923+2,515+1,421+ &147+277+333+   &8+8+8+ &0+0+8+ &2+2+1+ &0+0+8+\\
                                &\ persist pandas DataFrame                            &+1,198+1,413       &+414+1,437     &+8+3   &+0+3   &+2+1   &+0+3\\
                                &Assess data quality, examine descriptive statistics   &619+625+1,002      &388+475+439    &10+15+15 &0+0+0 &2+2+2 &10+15+0\\
                                &Utilize regular expressions                           &1,390              &263            &20 &0 &2 & 20\\
                                &\bf Project 8 Total &\bf 13,106    &\bf 4,173         &\bf 95 &\bf 11 (11.6\%) &\bf 16 &\bf 56 (58.9\%)\\
    \midrule
                                &\bf Course Total    &\bf 68,359     &\bf 37,713      &\bf 760 &\bf 407 (53.6\%) &\bf 126 &\bf 505 (66.4\%)\\
    \cline{2-8}
  \end{tabular}
\end{table*}

Table \ref{tab:ppp_concepts_results} reports the results of applying the models to the MCQs in PPP. The coding tasks results are presented in Table \ref{tab:project_results}. In PPP, learners are required to obtain at least 70\% from each unit and 75\% overall. The tests contribute 20\% towards the grade and projects the remaining 80\%. Learners additionally obtain 5 points per project for reflecting on their learning experience (text). It is fair to assume that GPT would easily generate passable reflections. Table \ref{tab:ppp_overall_results} shows that under this grading scheme GPT passed 4/ 8 course units, failing the course with the overall score of 67.3\%.%Hence, for example the model would pass Unit 1 of PPP with the score of 91.8\% ($[91+5]*0.8+75*0.2)$. Using this formula, the model successfully passed Unit 1 on foundations (91.8\%), Unit 3 on data structures (80.9\%), Unit 4 on object-oriented programming (89.4\%), and Unit 6 on data manipulation (87.7\%). It failed Unit 2 on control flow (56.1\%), Unit 5 on software development (25.2\%), Unit 7 on web scraping and office documents (50.4\%), and Unit 8 on data analysis (56.8\%). This amounts to failing the course with the overall score of 67.3\%.

\begin{table}
  \caption{PPP overall results. At least 70\% from each unit (projects and reflections contribute 80\%, quizzes 20\%) is required for passing. Course total needs to be >75.0\%. Green indicates passing and red failure.}
  \label{tab:ppp_overall_results}
  \footnotesize
  \begin{tabular}{lrrrr}
  \toprule
          & Project & Reflection & Test &\bf Unit  \\
  Unit    & \multicolumn{2}{c}{(weight 0.8)} & (w. 0.2) &\bf Total\\
  \midrule
  U1: Foundations                 & 91/95   & 5/5        & 9/12 &\cellcolor{green!15} 91.8\% \\
  U2: Control Flow                & 47/95   & 5/5        & 8/11 &\cellcolor{red!15} 56.1\% \\
  U3: Data Structures             & 80/95   & 5/5        & 9/14 &\cellcolor{green!15} 80.9\% \\
  U4: Object-oriented Programming & 84/95   & 5/5        & 10/11&\cellcolor{green!15} 89.4\% \\
  U5: Software Development        & 9/95    & 5/5        & 7/10 &\cellcolor{red!15} 25.2\% \\
  U6: Data Manipulation           & 95/95   & 5/5        & 5/13 &\cellcolor{green!15} 87.7\% \\
  U7: Web Scraping and Documents  & 43/95   & 5/5        & 3/5  &\cellcolor{red!15} 50.4\% \\
  U8: Data Analysis               & 56/95   & 5/5        & 2/5  &\cellcolor{red!15} 56.8\% \\
  \midrule
         &            &\multicolumn{2}{l}{\bf Course Total}&\cellcolor{red!15}\bf 67.3\% \\
  \cline{3-5}
  \end{tabular}
\end{table}

\section{Discussion}
\label{sec:discussion}
The results of applying \verb|text-davinci-003| to MCQ-style assessments suggest that there is no systematic difference between GPT's handling of introductory and intermediate topics. Contrary to what one might expect, the model passed some MCQ assessments related to more complex topics while failing the assessments of some introductory modules (1/5 modules passed in introductory PE1; 3/5 passed in intermediate PE2). % The results seem similar to Pan et al's common-sense science question answering task, with the PPP results falling in line with the USMLE/AICPA results as well, suggesting that GPT's handling of each of these topics are similar \cite{pan2022, kung2022performance, Gilson2022HowWD, Lievin2022CanLL}. 
%Some other properties of MCQs are likely decisive for GPT's handling of the question.
There appears to be a difference in success rate between MCQs that contain a code snippet (56.7\%) and those that do not (76.3\%). We hypothesize that either (a) a combination of natural language and a code snippet, and/or (b) a chain of reasoning steps are posing a challenge for GPT. This is observable on the difference in performance on PE1's quizzes (4/4 passed) compared to the performance on the module tests (1/4). In subsequent work, we analyzed the performance of the GPT models across MCQ categories to determine if questions of a certain type are handled more successfully than questions of other types \cite{savelka2023}.

%Test questions tended to require deeper more complex reasoning than quizzes. For example, while the quizzes often asked simple recall questions (e.g., ``What is the file extension for Python language files?''), the tests typically asked questions, such as ``What does \verb|1/1| evaluate to?''. In order to answer such a question, one even needs the correct understanding of topics not mentioned in the question (e.g., implicit type conversion).

%If a programming instructor would like to draft an MCQ-style assessment that would make it difficult for learners to use direct answers from a tool such as ChatGPT, it appears that one possible path might be to increase the ratio of questions assessing the skills at the \emph{apply}, \emph{analyze}, and \emph{evaluate} levels of Bloom's taxonomy~\cite{bloom1964taxonomy,conklin2005taxonomy} as opposed to the lower \emph{remember} and \emph{understand} levels. For example, questions involving small snippets of code requiring students to perform multiple reasoning steps could be successful in this regard. 

The results of generating solutions to coding tasks (Table \ref{tab:project_results}) offer a similar insight as the results from answering the MCQs in that it appears to make little difference whether a topic is introductory or intermediate (e.g., 49.5\% on control flow vs 88.4\% on OOP activities). It rather appears that the level of depth in which the coding activities are exposing the topic to a learner matters when it comes to GPT's ability to generate a correct solution. For example, the OOP activities focus on the use of foundational OOP constructs (e.g.,~defining a class, implementing a constructor). The control flow activities are more in-depth as the learner is expected to handle fine-grained details of \verb|if-elif-else| statements, \verb|for| and \verb|while| loops, as well as combining them into more complex scripts.

We observed that coding tasks involving nuanced output formatting (e.g., a tabular report to be printed in a terminal), bug-fixing~(e.g., a typical ``off-by-one'' bug), refactoring (e.g., fix style issues based on the output of style-checking code), and/or real-world artifacts (e.g., an input \verb|CSV| file with column names that differ from the names in the desired \verb|DataFrame|) posed the greatest challenge to GPT. This appears to be consistent with our findings related to the causes of difficulties in handling MCQ exercises (i.e., combining multiple different artifacts, requirement for a chain of reasoning) as well as with the findings reported in existing work \cite{10.1145/3511861.3511863,wermelinger2023using,https://doi.org/10.48550/arxiv.2107.03374}. % Increasing the use of coding tasks with these properties might be desirable if a programming instructor would aim to minimize the chances of learners being able to rely on tools such as Copilot for end-to-end solutions to their programming assignments.
Programming instructors might consider increasing the use of coding tasks with these properties, if they wish to discourage learners' use of tools like Copilot for end-to-end solutions.

\begin{figure}
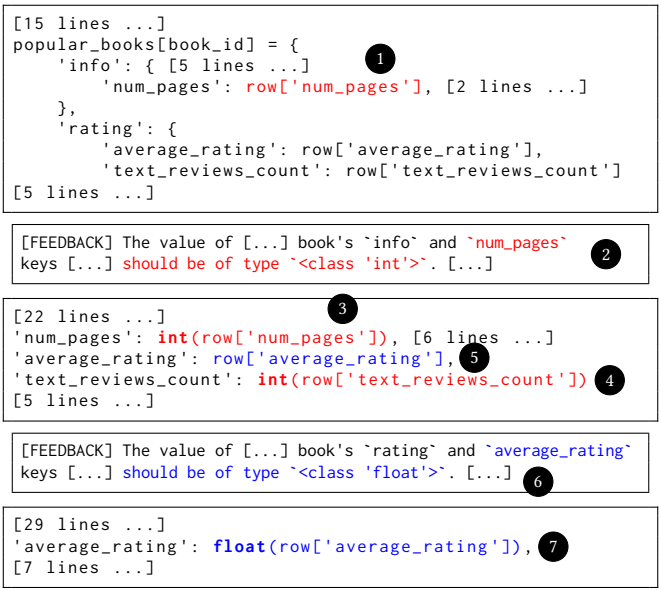

\footnotesize
\begin{lstlisting}[frame=single,style=base]
[15 lines ...]
popular_books[book_id] = {
    'info': { [5 lines ...]
        'num_pages': @row['num_pages']@, [2 lines ...]
    },
    'rating': {
        'average_rating': row['average_rating'],
        'text_reviews_count': row['text_reviews_count']
[5 lines ...]
\end{lstlisting}
\begin{textblock*}{1.4cm}(4.2cm,-2.35cm)
\circled{1}
\end{textblock*}

\begin{Verbatim}[frame=single,commandchars=\\\{\}]
[FEEDBACK] The value of [...] book's `info` and \textcolor{red}{`num_pages`} 
keys [...] \textcolor{red}{should be of type `<class 'int'>`}. [...]
\end{Verbatim}
\begin{textblock*}{1.4cm}(7.2cm,-0.65cm)
\circled{2}
\end{textblock*}

\begin{lstlisting}[frame=single,style=base]
[22 lines ...]
'num_pages': @int(row['num_pages'])@, [6 lines ...]
'average_rating': |row['average_rating']|,
'text_reviews_count': @int(row['text_reviews_count'])@
[5 lines ...]
\end{lstlisting}
\begin{textblock*}{1.4cm}(3.7cm,-1.8cm)
\circled{3}
\end{textblock*}
\begin{textblock*}{1.4cm}(7.25cm,-0.85cm)
\circled{4}
\end{textblock*}
\begin{textblock*}{1.4cm}(5.45cm,-1.15cm)
\circled{5}
\end{textblock*}

\begin{Verbatim}[frame=single,commandchars=\\\{\}]
[FEEDBACK] The value of [...] book's `rating` and \textcolor{blue}{`average_rating`} 
keys [...] \textcolor{blue}{should be of type `<class 'float'>`}. [...]
\end{Verbatim}
\begin{textblock*}{1.4cm}(6.3cm,-0.4cm)
\circled{6}
\end{textblock*}

\begin{lstlisting}[frame=single,style=base]
[29 lines ...]
'average_rating': |float(row['average_rating'])|, 
[7 lines ...]
\end{lstlisting}
\begin{textblock*}{1.4cm}(6.5cm,-0.85cm)
\circled{7}
\end{textblock*}
\caption{Incorrect GPT's solution is evaluated by an auto-grader which recognizes that the num\_pages field (1) should not be of type str and produces corresponding feedback (2). GPT corrects the flaw (3) as well as similar one (4) not mentioned in the feedback. An issue with the average\_rating field~(5) is fixed (7) based on the additional feedback (6).}
\label{fig:example-success}
\end{figure}

\begin{figure}
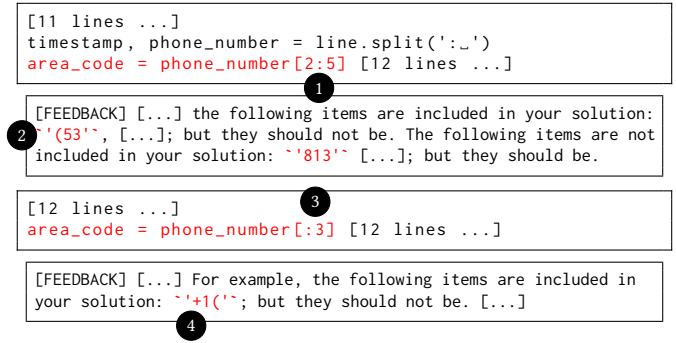

\footnotesize
\begin{lstlisting}[frame=single,style=base]
[11 lines ...]
timestamp, phone_number = line.split(': ')
@area_code = phone_number[2:5]@ [12 lines ...]
\end{lstlisting}
\begin{textblock*}{1.4cm}(3.2cm,-0.25cm)
\circled{1}
\end{textblock*}

\begin{Verbatim}[frame=single,commandchars=\\\{\}]
[FEEDBACK] [...] the following items are included in your solution: 
\textcolor{red}{`'(53'`}, [...]; but they should not be. The following items are not 
included in your solution: \textcolor{red}{`'813'`} [...]; but they should be.
\end{Verbatim}
\begin{textblock*}{1.4cm}(-0.75cm,-0.8cm)
\circled{2}
\end{textblock*}

\begin{lstlisting}[frame=single,style=base]
[12 lines ...]
@area_code = phone_number[:3]@ [12 lines ...]
\end{lstlisting}
\begin{textblock*}{1.4cm}(3.15cm,-0.95cm)
\circled{3}
\end{textblock*}

\begin{Verbatim}[frame=single,commandchars=\\\{\}]
[FEEDBACK] [...] For example, the following items are included in 
your solution: \textcolor{red}{`'+1('`}; but they should not be. [...]
\end{Verbatim}
\begin{textblock*}{1.4cm}(1.5cm,-0.2cm)
\circled{4}
\end{textblock*}

% \begin{lstlisting}[frame=single,style=base]
% [12 lines ...]
% @area_code = phone_number[:3]@
% [12 lines ...]
% \end{lstlisting}
\caption{Incorrect GPT's solution (1) is evaluated by an auto-grader which recognizes that the extracted area code (2) is not as expected. Instead of modifying the str slice indices to [3:6] which would fixed the issue they are modified to [:3]~(3). This again produces an incorrect output (4).}
\label{fig:example-fail}
\end{figure}

%\TODO{RQ5 Feedback handling}
The results in Table \ref{tab:ppp_overall_results} also show that GPT is able to successfully utilize auto-grader feedback primarily designed for a human learner. The overall projects' score of 53.6\% from the first submissions increased to 66.4\% due to revised submissions based on the feedback. Figure~\ref{fig:example-success} shows an example of GPT's successful handling of the provided feedback. In multiple steps, a faulty solution was transformed into a correct one. Specifically, code generating a \verb|JSON| file is iteratively amended to coerce \verb|str| fields to the expected types (\verb|int| and \verb|float|). On the other hand, Figure \ref{fig:example-fail} provides an example of GPT not utilizing the feedback well. Here, an expected output compared to the actual one does not lead GPT to modify the string slicing code as needed. The observed trend illustrated by the two examples is consistent with other observations presented in this paper. A feedback that directly points to the problem is typically utilized well. Whereas feedback that requires a chain of reasoning (e.g.,~actual vs. expected output) is much more challenging for GPT.

\section{Conclusions and Future Work}
We evaluated \verb|text-davinci-003| on a sizeable set of 599 diverse assessments (MCQs, coding tasks) from three Python programming courses. The model's answers to MCQs and its solutions to coding tasks obtained non-trivial portions of the available scores (56.1\%, 67.9\%, and 67.3\%). While these are not sufficient for successful completion of the courses, the model exhibits remarkable capabilities, including the ability to correct solutions based on auto-grader feedback. The most important limitation of the  GPT models is their apparent struggle with activities that require chains of reasoning steps (e.g., non-trivial reasoning about a code snippet or fixing a script based on the actual vs. expected output). Our findings are important for programming instructors who intend to adapt their classes to account for the existence of the powerful QA and code generating technology. As the tools and the underlying models evolve, it appears that in future programming education code authoring may need to be augmented with emphasis on requirements formulation, debugging, trade-off analysis, and critical thinking.

While our study of GPT's performance on diverse types of programming assessments yielded numerous valuable insights, it is subject to countless limitations and leaves much room for improvement. Hence, we suggest several directions for future work: (i)~further analyze the effects of prompt-tuning (ii) and/or iterative prompt-construction; (iii) examine the poor performance of GPT on MCQs containing code snippets; (iv) get deeper insight into the properties of questions and coding tasks that are challenging for GPT; and (v)~study possibilities of effective integration of GPT-based tools, e.g., ChatGPT or Copilot, into programming education.

% Draft - Arav. Likely needs some revisions.
%Considering our results, future work is needed to extend and continue to understand the potential for GPT models in CS1 contexts. Given the possibility for academic misuse of these tools, there seems to be potential in using GPT models to assess question difficulty ahead of giving an examination, which could reduce the amount of time needed to provide quality assurance for new quizzes. With the release of ChatGPT's API in the future, we expect to be able to do far more extensive comparisons between current models, which should drive even more research. Lastly, we expect future work to be focused on prompt-tuning, in particular to understand how hard is it for students to tune prompts accordingly for GPT models to solve tasks, and to understand systematically what kinds of prompt-tuning are necessary to get GPT models to perform better on these exercises.

% Potential usage of GPT as a difficulty measurer??

% \begin{acks}
% Lorem ipsum
% \end{acks}

%%
%% The next two lines define the bibliography style to be used, and
%% the bibliography file.
\bibliographystyle{ACM-Reference-Format}
\bibliography{sample-base}

%%% -*-BibTeX-*-
%%% Do NOT edit. File created by BibTeX with style
%%% ACM-Reference-Format-Journals [18-Jan-2012].

\begin{thebibliography}{38}

%%% ====================================================================
%%% NOTE TO THE USER: you can override these defaults by providing
%%% customized versions of any of these macros before the \bibliography
%%% command.  Each of them MUST provide its own final punctuation,
%%% except for \shownote{}, \showDOI{}, and \showURL{}.  The latter two
%%% do not use final punctuation, in order to avoid confusing it with
%%% the Web address.
%%%
%%% To suppress output of a particular field, define its macro to expand
%%% to an empty string, or better, \unskip, like this:
%%%
%%% \newcommand{\showDOI}[1]{\unskip}   % LaTeX syntax
%%%
%%% \def \showDOI #1{\unskip}           % plain TeX syntax
%%%
%%% ====================================================================

\ifx \showCODEN    \undefined \def \showCODEN     #1{\unskip}     \fi
\ifx \showDOI      \undefined \def \showDOI       #1{#1}\fi
\ifx \showISBNx    \undefined \def \showISBNx     #1{\unskip}     \fi
\ifx \showISBNxiii \undefined \def \showISBNxiii  #1{\unskip}     \fi
\ifx \showISSN     \undefined \def \showISSN      #1{\unskip}     \fi
\ifx \showLCCN     \undefined \def \showLCCN      #1{\unskip}     \fi
\ifx \shownote     \undefined \def \shownote      #1{#1}          \fi
\ifx \showarticletitle \undefined \def \showarticletitle #1{#1}   \fi
\ifx \showURL      \undefined \def \showURL       {\relax}        \fi
% The following commands are used for tagged output and should be
% invisible to TeX
\providecommand\bibfield[2]{#2}
\providecommand\bibinfo[2]{#2}
\providecommand\natexlab[1]{#1}
\providecommand\showeprint[2][]{arXiv:#2}

\bibitem[Ankur and Atul(2022)]%
        {ankur2022}
\bibfield{author}{\bibinfo{person}{Desai Ankur} {and} \bibinfo{person}{Deo
  Atul}.} \bibinfo{year}{2022}\natexlab{}.
\newblock \showarticletitle{Introducing Amazon CodeWhisperer, the ML-powered
  Coding Companion}.
\newblock \bibinfo{journal}{\emph{AWS Machine Learning Blog}}
  (\bibinfo{year}{2022}).
\newblock
Issue June 24, 2022.
\urldef\tempurl%
\url{https://aws.amazon.com/blogs/machine-learning/introducing-amazon-codewhisperer-the-ml-powered-coding-companion/}
\showURL{%
\tempurl}


\bibitem[Barke et~al\mbox{.}(2022)]%
        {Barke2022GroundedCH}
\bibfield{author}{\bibinfo{person}{Shraddha Barke}, \bibinfo{person}{Michael~B.
  James}, {and} \bibinfo{person}{Nadia Polikarpova}.}
  \bibinfo{year}{2022}\natexlab{}.
\newblock \showarticletitle{Grounded Copilot: How Programmers Interact with
  Code-Generating Models}.
\newblock \bibinfo{journal}{\emph{ArXiv}}  \bibinfo{volume}{abs/2206.15000}
  (\bibinfo{year}{2022}).
\newblock


\bibitem[Becker et~al\mbox{.}(2022)]%
        {becker2022programming}
\bibfield{author}{\bibinfo{person}{Brett~A Becker}, \bibinfo{person}{Paul
  Denny}, \bibinfo{person}{James Finnie-Ansley}, \bibinfo{person}{Andrew
  Luxton-Reilly}, \bibinfo{person}{James Prather}, {and}
  \bibinfo{person}{Eddie~Antonio Santos}.} \bibinfo{year}{2022}\natexlab{}.
\newblock \showarticletitle{Programming Is Hard--Or at Least It Used to Be:
  Educational Opportunities And Challenges of AI Code Generation}.
\newblock \bibinfo{journal}{\emph{arXiv preprint arXiv:2212.01020}}
  (\bibinfo{year}{2022}).
\newblock


\bibitem[Biderman and Raff(2022)]%
        {Biderman2022FoolingMD}
\bibfield{author}{\bibinfo{person}{Stella~Rose Biderman} {and}
  \bibinfo{person}{Edward Raff}.} \bibinfo{year}{2022}\natexlab{}.
\newblock \showarticletitle{Fooling MOSS Detection with Pretrained Language
  Models}.
\newblock \bibinfo{journal}{\emph{Proceedings of the 31st ACM International
  Conference on Information \& Knowledge Management}} (\bibinfo{year}{2022}).
\newblock


\bibitem[Bommarito et~al\mbox{.}(2023)]%
        {bommarito2023gpt}
\bibfield{author}{\bibinfo{person}{Jillian Bommarito}, \bibinfo{person}{Michael
  Bommarito}, \bibinfo{person}{Daniel~Martin Katz}, {and}
  \bibinfo{person}{Jessica Katz}.} \bibinfo{year}{2023}\natexlab{}.
\newblock \showarticletitle{GPT as Knowledge Worker: A Zero-Shot Evaluation of
  (AI) CPA Capabilities}.
\newblock \bibinfo{journal}{\emph{arXiv preprint arXiv:2301.04408}}
  (\bibinfo{year}{2023}).
\newblock


\bibitem[Bommarito~II and Katz(2022)]%
        {bommarito2022gpt}
\bibfield{author}{\bibinfo{person}{Michael Bommarito~II} {and}
  \bibinfo{person}{Daniel~Martin Katz}.} \bibinfo{year}{2022}\natexlab{}.
\newblock \showarticletitle{GPT Takes the Bar Exam}.
\newblock \bibinfo{journal}{\emph{arXiv preprint arXiv:2212.14402}}
  (\bibinfo{year}{2022}).
\newblock


\bibitem[Bowman(2023)]%
        {Bowman2023}
\bibfield{author}{\bibinfo{person}{Emma Bowman}.}
  \bibinfo{year}{2023}\natexlab{}.
\newblock \showarticletitle{A college student created an app that can tell
  whether AI wrote an essay}.
\newblock \bibinfo{journal}{\emph{NPR Technology}} (\bibinfo{year}{2023}).
\newblock
Issue January 9, 2023.
\urldef\tempurl%
\url{https://www.npr.org/2023/01/09/1147549845/gptzero-ai-chatgpt-edward-tian-plagiarism}
\showURL{%
\tempurl}


\bibitem[Brown et~al\mbox{.}(2020)]%
        {brown2020language}
\bibfield{author}{\bibinfo{person}{Tom Brown}, \bibinfo{person}{Benjamin Mann},
  \bibinfo{person}{Nick Ryder}, \bibinfo{person}{Melanie Subbiah},
  \bibinfo{person}{Jared~D Kaplan}, \bibinfo{person}{Prafulla Dhariwal},
  \bibinfo{person}{Arvind Neelakantan}, \bibinfo{person}{Pranav Shyam},
  \bibinfo{person}{Girish Sastry}, \bibinfo{person}{Amanda Askell},
  {et~al\mbox{.}}} \bibinfo{year}{2020}\natexlab{}.
\newblock \showarticletitle{Language models are few-shot learners}.
\newblock \bibinfo{journal}{\emph{Advances in neural information processing
  systems}}  \bibinfo{volume}{33} (\bibinfo{year}{2020}),
  \bibinfo{pages}{1877--1901}.
\newblock


\bibitem[Chen et~al\mbox{.}(2021)]%
        {https://doi.org/10.48550/arxiv.2107.03374}
\bibfield{author}{\bibinfo{person}{Mark Chen}, \bibinfo{person}{Jerry Tworek},
  \bibinfo{person}{Heewoo Jun}, \bibinfo{person}{Qiming Yuan},
  \bibinfo{person}{Henrique Ponde de~Oliveira Pinto}, \bibinfo{person}{Jared
  Kaplan}, \bibinfo{person}{Harri Edwards}, \bibinfo{person}{Yuri Burda},
  \bibinfo{person}{Nicholas Joseph}, \bibinfo{person}{Greg Brockman},
  \bibinfo{person}{Alex Ray}, \bibinfo{person}{Raul Puri},
  \bibinfo{person}{Gretchen Krueger}, \bibinfo{person}{Michael Petrov},
  \bibinfo{person}{Heidy Khlaaf}, \bibinfo{person}{Girish Sastry},
  \bibinfo{person}{Pamela Mishkin}, \bibinfo{person}{Brooke Chan},
  \bibinfo{person}{Scott Gray}, \bibinfo{person}{Nick Ryder},
  \bibinfo{person}{Mikhail Pavlov}, \bibinfo{person}{Alethea Power},
  \bibinfo{person}{Lukasz Kaiser}, \bibinfo{person}{Mohammad Bavarian},
  \bibinfo{person}{Clemens Winter}, \bibinfo{person}{Philippe Tillet},
  \bibinfo{person}{Felipe~Petroski Such}, \bibinfo{person}{Dave Cummings},
  \bibinfo{person}{Matthias Plappert}, \bibinfo{person}{Fotios Chantzis},
  \bibinfo{person}{Elizabeth Barnes}, \bibinfo{person}{Ariel Herbert-Voss},
  \bibinfo{person}{William~Hebgen Guss}, \bibinfo{person}{Alex Nichol},
  \bibinfo{person}{Alex Paino}, \bibinfo{person}{Nikolas Tezak},
  \bibinfo{person}{Jie Tang}, \bibinfo{person}{Igor Babuschkin},
  \bibinfo{person}{Suchir Balaji}, \bibinfo{person}{Shantanu Jain},
  \bibinfo{person}{William Saunders}, \bibinfo{person}{Christopher Hesse},
  \bibinfo{person}{Andrew~N. Carr}, \bibinfo{person}{Jan Leike},
  \bibinfo{person}{Josh Achiam}, \bibinfo{person}{Vedant Misra},
  \bibinfo{person}{Evan Morikawa}, \bibinfo{person}{Alec Radford},
  \bibinfo{person}{Matthew Knight}, \bibinfo{person}{Miles Brundage},
  \bibinfo{person}{Mira Murati}, \bibinfo{person}{Katie Mayer},
  \bibinfo{person}{Peter Welinder}, \bibinfo{person}{Bob McGrew},
  \bibinfo{person}{Dario Amodei}, \bibinfo{person}{Sam McCandlish},
  \bibinfo{person}{Ilya Sutskever}, {and} \bibinfo{person}{Wojciech Zaremba}.}
  \bibinfo{year}{2021}\natexlab{}.
\newblock \bibinfo{title}{Evaluating Large Language Models Trained on Code}.
\newblock
\newblock
\urldef\tempurl%
\url{https://doi.org/10.48550/ARXIV.2107.03374}
\showDOI{\tempurl}


\bibitem[Denny et~al\mbox{.}(2022)]%
        {https://doi.org/10.48550/arxiv.2210.15157}
\bibfield{author}{\bibinfo{person}{Paul Denny}, \bibinfo{person}{Viraj Kumar},
  {and} \bibinfo{person}{Nasser Giacaman}.} \bibinfo{year}{2022}\natexlab{}.
\newblock \bibinfo{title}{Conversing with Copilot: Exploring Prompt Engineering
  for Solving CS1 Problems Using Natural Language}.
\newblock
\newblock
\urldef\tempurl%
\url{https://doi.org/10.48550/ARXIV.2210.15157}
\showDOI{\tempurl}


\bibitem[Drori and Verma(2021)]%
        {https://doi.org/10.48550/arxiv.2111.08171}
\bibfield{author}{\bibinfo{person}{Iddo Drori} {and} \bibinfo{person}{Nakul
  Verma}.} \bibinfo{year}{2021}\natexlab{}.
\newblock \bibinfo{title}{Solving Linear Algebra by Program Synthesis}.
\newblock
\newblock
\urldef\tempurl%
\url{https://doi.org/10.48550/ARXIV.2111.08171}
\showDOI{\tempurl}


\bibitem[Elsen-Rooney(2023)]%
        {ElsenRooney2023}
\bibfield{author}{\bibinfo{person}{Michael Elsen-Rooney}.}
  \bibinfo{year}{2023}\natexlab{}.
\newblock \showarticletitle{NYC education department blocks ChatGPT on school
  devices, networks}.
\newblock \bibinfo{journal}{\emph{Chalkbeat New York}} (\bibinfo{year}{2023}).
\newblock
Issue January 3, 2023.
\urldef\tempurl%
\url{https://ny.chalkbeat.org/2023/1/3/23537987/nyc-schools-ban-chatgpt-writing-artificial-intelligence}
\showURL{%
\tempurl}


\bibitem[Finnie-Ansley et~al\mbox{.}(2022)]%
        {10.1145/3511861.3511863}
\bibfield{author}{\bibinfo{person}{James Finnie-Ansley}, \bibinfo{person}{Paul
  Denny}, \bibinfo{person}{Brett~A. Becker}, \bibinfo{person}{Andrew
  Luxton-Reilly}, {and} \bibinfo{person}{James Prather}.}
  \bibinfo{year}{2022}\natexlab{}.
\newblock \showarticletitle{The Robots Are Coming: Exploring the Implications
  of OpenAI Codex on Introductory Programming}. In
  \bibinfo{booktitle}{\emph{Australasian Computing Education Conference}}
  (Virtual Event, Australia) \emph{(\bibinfo{series}{ACE '22})}.
  \bibinfo{publisher}{Association for Computing Machinery},
  \bibinfo{address}{New York, NY, USA}, \bibinfo{pages}{10–19}.
\newblock
\showISBNx{9781450396431}
\urldef\tempurl%
\url{https://doi.org/10.1145/3511861.3511863}
\showDOI{\tempurl}


\bibitem[Finnie-Ansley et~al\mbox{.}(2023)]%
        {finnie2023my}
\bibfield{author}{\bibinfo{person}{James Finnie-Ansley}, \bibinfo{person}{Paul
  Denny}, \bibinfo{person}{Andrew Luxton-Reilly},
  \bibinfo{person}{Eddie~Antonio Santos}, \bibinfo{person}{James Prather},
  {and} \bibinfo{person}{Brett~A Becker}.} \bibinfo{year}{2023}\natexlab{}.
\newblock \showarticletitle{My AI Wants to Know if This Will Be on the Exam:
  Testing OpenAI’s Codex on CS2 Programming Exercises}. In
  \bibinfo{booktitle}{\emph{Proceedings of the 25th Australasian Computing
  Education Conference}}. \bibinfo{pages}{97--104}.
\newblock


\bibitem[Gilson et~al\mbox{.}(2022)]%
        {Gilson2022HowWD}
\bibfield{author}{\bibinfo{person}{Amelia Gilson}, \bibinfo{person}{Conrad~W.
  Safranek}, \bibinfo{person}{Tao Huang}, \bibinfo{person}{Vimig Socrates},
  \bibinfo{person}{Lim~Sze Chi}, \bibinfo{person}{Roderick~A. Taylor}, {and}
  \bibinfo{person}{David Chartash}.} \bibinfo{year}{2022}\natexlab{}.
\newblock \showarticletitle{How Well Does ChatGPT Do When Taking the Medical
  Licensing Exams? The Implications of Large Language Models for Medical
  Education and Knowledge Assessment}. In \bibinfo{booktitle}{\emph{medRxiv}}.
\newblock


\bibitem[Hendrycks et~al\mbox{.}(2020)]%
        {hendrycks2022}
\bibfield{author}{\bibinfo{person}{Dan Hendrycks}, \bibinfo{person}{Collin
  Burns}, \bibinfo{person}{Steven Basart}, \bibinfo{person}{Andy Zou},
  \bibinfo{person}{Mantas Mazeika}, \bibinfo{person}{Dawn Song}, {and}
  \bibinfo{person}{Jacob Steinhardt}.} \bibinfo{year}{2020}\natexlab{}.
\newblock \bibinfo{title}{Measuring Massive Multitask Language Understanding}.
\newblock
\newblock
\urldef\tempurl%
\url{https://doi.org/10.48550/ARXIV.2009.03300}
\showDOI{\tempurl}


\bibitem[Huang(2023)]%
        {Huang2023}
\bibfield{author}{\bibinfo{person}{Kalley Huang}.}
  \bibinfo{year}{2023}\natexlab{}.
\newblock \showarticletitle{Alarmed by A.I. Chatbots, Universities Start
  Revamping How They Teach}.
\newblock \bibinfo{journal}{\emph{New York Times}} (\bibinfo{year}{2023}).
\newblock
Issue January 16, 2023.
\urldef\tempurl%
\url{https://www.nytimes.com/2023/01/16/technology/chatgpt-artificial-intelligence-universities.html}
\showURL{%
\tempurl}


\bibitem[Jalil et~al\mbox{.}(2023)]%
        {jalil2023chatgpt}
\bibfield{author}{\bibinfo{person}{Sajed Jalil}, \bibinfo{person}{Suzzana
  Rafi}, \bibinfo{person}{Thomas~D LaToza}, \bibinfo{person}{Kevin Moran},
  {and} \bibinfo{person}{Wing Lam}.} \bibinfo{year}{2023}\natexlab{}.
\newblock \showarticletitle{ChatGPT and Software Testing Education: Promises \&
  Perils}.
\newblock \bibinfo{journal}{\emph{arXiv preprint arXiv:2302.03287}}
  (\bibinfo{year}{2023}).
\newblock


\bibitem[Karmakar et~al\mbox{.}(2022)]%
        {Karmakar2022CodexHH}
\bibfield{author}{\bibinfo{person}{Anjan Karmakar},
  \bibinfo{person}{Julian~Aron Prenner}, \bibinfo{person}{Marco D'Ambros},
  {and} \bibinfo{person}{Romain Robbes}.} \bibinfo{year}{2022}\natexlab{}.
\newblock \showarticletitle{Codex Hacks HackerRank: Memorization Issues and a
  Framework for Code Synthesis Evaluation}.
\newblock \bibinfo{journal}{\emph{ArXiv}}  \bibinfo{volume}{abs/2212.02684}
  (\bibinfo{year}{2022}).
\newblock


\bibitem[Karnalim et~al\mbox{.}(2019)]%
        {karnalim2019source}
\bibfield{author}{\bibinfo{person}{Oscar Karnalim}, \bibinfo{person}{Setia
  Budi}, \bibinfo{person}{Hapnes Toba}, {and} \bibinfo{person}{Mike Joy}.}
  \bibinfo{year}{2019}\natexlab{}.
\newblock \showarticletitle{Source Code Plagiarism Detection in Academia with
  Information Retrieval: Dataset and the Observation.}
\newblock \bibinfo{journal}{\emph{Informatics in Education}}
  \bibinfo{volume}{18}, \bibinfo{number}{2} (\bibinfo{year}{2019}),
  \bibinfo{pages}{321--344}.
\newblock


\bibitem[Kokotsaki et~al\mbox{.}(2016)]%
        {kokotsaki2016project}
\bibfield{author}{\bibinfo{person}{Dimitra Kokotsaki},
  \bibinfo{person}{Victoria Menzies}, {and} \bibinfo{person}{Andy Wiggins}.}
  \bibinfo{year}{2016}\natexlab{}.
\newblock \showarticletitle{Project-based learning: A review of the
  literature}.
\newblock \bibinfo{journal}{\emph{Improving schools}} \bibinfo{volume}{19},
  \bibinfo{number}{3} (\bibinfo{year}{2016}), \bibinfo{pages}{267--277}.
\newblock


\bibitem[Kung et~al\mbox{.}(2022)]%
        {kung2022performance}
\bibfield{author}{\bibinfo{person}{Tiffany~H Kung}, \bibinfo{person}{Morgan
  Cheatham}, \bibinfo{person}{Arielle Medinilla}, \bibinfo{person}{Czarina
  Sillos}, \bibinfo{person}{Lorie De~Leon}, \bibinfo{person}{Camille Elepano},
  \bibinfo{person}{Marie Madriaga}, \bibinfo{person}{Rimel Aggabao},
  \bibinfo{person}{Giezel Diaz-Candido}, \bibinfo{person}{James Maningo},
  {et~al\mbox{.}}} \bibinfo{year}{2022}\natexlab{}.
\newblock \showarticletitle{Performance of ChatGPT on USMLE: Potential for
  AI-Assisted Medical Education Using Large Language Models}.
\newblock \bibinfo{journal}{\emph{medRxiv}} (\bibinfo{year}{2022}).
\newblock


\bibitem[Lai et~al\mbox{.}(2017)]%
        {lai2017race}
\bibfield{author}{\bibinfo{person}{Guokun Lai}, \bibinfo{person}{Qizhe Xie},
  \bibinfo{person}{Hanxiao Liu}, \bibinfo{person}{Yiming Yang}, {and}
  \bibinfo{person}{Eduard Hovy}.} \bibinfo{year}{2017}\natexlab{}.
\newblock \showarticletitle{Race: Large-scale reading comprehension dataset
  from examinations}.
\newblock \bibinfo{journal}{\emph{arXiv preprint arXiv:1704.04683}}
  (\bibinfo{year}{2017}).
\newblock


\bibitem[Li et~al\mbox{.}(2022)]%
        {doi:10.1126/science.abq1158}
\bibfield{author}{\bibinfo{person}{Yujia Li}, \bibinfo{person}{David Choi},
  \bibinfo{person}{Junyoung Chung}, \bibinfo{person}{Nate Kushman},
  \bibinfo{person}{Julian Schrittwieser}, \bibinfo{person}{Rémi Leblond},
  \bibinfo{person}{Tom Eccles}, \bibinfo{person}{James Keeling},
  \bibinfo{person}{Felix Gimeno}, \bibinfo{person}{Agustin~Dal Lago},
  \bibinfo{person}{Thomas Hubert}, \bibinfo{person}{Peter Choy},
  \bibinfo{person}{Cyprien de Masson~d’Autume}, \bibinfo{person}{Igor
  Babuschkin}, \bibinfo{person}{Xinyun Chen}, \bibinfo{person}{Po-Sen Huang},
  \bibinfo{person}{Johannes Welbl}, \bibinfo{person}{Sven Gowal},
  \bibinfo{person}{Alexey Cherepanov}, \bibinfo{person}{James Molloy},
  \bibinfo{person}{Daniel~J. Mankowitz}, \bibinfo{person}{Esme~Sutherland
  Robson}, \bibinfo{person}{Pushmeet Kohli}, \bibinfo{person}{Nando de
  Freitas}, \bibinfo{person}{Koray Kavukcuoglu}, {and} \bibinfo{person}{Oriol
  Vinyals}.} \bibinfo{year}{2022}\natexlab{}.
\newblock \showarticletitle{Competition-level code generation with AlphaCode}.
\newblock \bibinfo{journal}{\emph{Science}} \bibinfo{volume}{378},
  \bibinfo{number}{6624} (\bibinfo{year}{2022}), \bibinfo{pages}{1092--1097}.
\newblock
\urldef\tempurl%
\url{https://doi.org/10.1126/science.abq1158}
\showDOI{\tempurl}
\showeprint{https://www.science.org/doi/pdf/10.1126/science.abq1158}


\bibitem[Li'evin et~al\mbox{.}(2022)]%
        {Lievin2022CanLL}
\bibfield{author}{\bibinfo{person}{Valentin Li'evin},
  \bibinfo{person}{Christoffer~Egeberg Hother}, {and} \bibinfo{person}{Ole
  Winther}.} \bibinfo{year}{2022}\natexlab{}.
\newblock \showarticletitle{Can large language models reason about medical
  questions?}
\newblock \bibinfo{journal}{\emph{ArXiv}}  \bibinfo{volume}{abs/2207.08143}
  (\bibinfo{year}{2022}).
\newblock


\bibitem[Lu et~al\mbox{.}(2022)]%
        {pan2022}
\bibfield{author}{\bibinfo{person}{Pan Lu}, \bibinfo{person}{Swaroop Mishra},
  \bibinfo{person}{Tony Xia}, \bibinfo{person}{Liang Qiu},
  \bibinfo{person}{Kai-Wei Chang}, \bibinfo{person}{Song-Chun Zhu},
  \bibinfo{person}{Oyvind Tafjord}, \bibinfo{person}{Peter Clark}, {and}
  \bibinfo{person}{Ashwin Kalyan}.} \bibinfo{year}{2022}\natexlab{}.
\newblock \bibinfo{title}{Learn to Explain: Multimodal Reasoning via Thought
  Chains for Science Question Answering}.
\newblock
\newblock
\urldef\tempurl%
\url{https://doi.org/10.48550/ARXIV.2209.09513}
\showDOI{\tempurl}


\bibitem[Mihaylov et~al\mbox{.}(2018)]%
        {mihaylov2018can}
\bibfield{author}{\bibinfo{person}{Todor Mihaylov}, \bibinfo{person}{Peter
  Clark}, \bibinfo{person}{Tushar Khot}, {and} \bibinfo{person}{Ashish
  Sabharwal}.} \bibinfo{year}{2018}\natexlab{}.
\newblock \showarticletitle{Can a suit of armor conduct electricity? a new
  dataset for open book question answering}.
\newblock \bibinfo{journal}{\emph{arXiv preprint arXiv:1809.02789}}
  (\bibinfo{year}{2018}).
\newblock


\bibitem[Mostafazadeh et~al\mbox{.}(2016)]%
        {mostafazadeh2016corpus}
\bibfield{author}{\bibinfo{person}{Nasrin Mostafazadeh},
  \bibinfo{person}{Nathanael Chambers}, \bibinfo{person}{Xiaodong He},
  \bibinfo{person}{Devi Parikh}, \bibinfo{person}{Dhruv Batra},
  \bibinfo{person}{Lucy Vanderwende}, \bibinfo{person}{Pushmeet Kohli}, {and}
  \bibinfo{person}{James Allen}.} \bibinfo{year}{2016}\natexlab{}.
\newblock \showarticletitle{A corpus and cloze evaluation for deeper
  understanding of commonsense stories}. In
  \bibinfo{booktitle}{\emph{Proceedings of the 2016 Conference of the North
  American Chapter of the Association for Computational Linguistics: Human
  Language Technologies}}. \bibinfo{pages}{839--849}.
\newblock


\bibitem[Nguyen and Nadi(2022)]%
        {9796235}
\bibfield{author}{\bibinfo{person}{Nhan Nguyen} {and} \bibinfo{person}{Sarah
  Nadi}.} \bibinfo{year}{2022}\natexlab{}.
\newblock \showarticletitle{An Empirical Evaluation of GitHub Copilot's Code
  Suggestions}. In \bibinfo{booktitle}{\emph{2022 IEEE/ACM 19th International
  Conference on Mining Software Repositories (MSR)}}. \bibinfo{pages}{1--5}.
\newblock
\urldef\tempurl%
\url{https://doi.org/10.1145/3524842.3528470}
\showDOI{\tempurl}


\bibitem[Ouyang et~al\mbox{.}(2022)]%
        {ouyang2022training}
\bibfield{author}{\bibinfo{person}{Long Ouyang}, \bibinfo{person}{Jeff Wu},
  \bibinfo{person}{Xu Jiang}, \bibinfo{person}{Diogo Almeida},
  \bibinfo{person}{Carroll~L Wainwright}, \bibinfo{person}{Pamela Mishkin},
  \bibinfo{person}{Chong Zhang}, \bibinfo{person}{Sandhini Agarwal},
  \bibinfo{person}{Katarina Slama}, \bibinfo{person}{Alex Ray},
  {et~al\mbox{.}}} \bibinfo{year}{2022}\natexlab{}.
\newblock \showarticletitle{Training language models to follow instructions
  with human feedback}.
\newblock \bibinfo{journal}{\emph{arXiv preprint arXiv:2203.02155}}
  (\bibinfo{year}{2022}).
\newblock


\bibitem[Pearce et~al\mbox{.}(2022)]%
        {pearce2022asleep}
\bibfield{author}{\bibinfo{person}{Hammond Pearce}, \bibinfo{person}{Baleegh
  Ahmad}, \bibinfo{person}{Benjamin Tan}, \bibinfo{person}{Brendan
  Dolan-Gavitt}, {and} \bibinfo{person}{Ramesh Karri}.}
  \bibinfo{year}{2022}\natexlab{}.
\newblock \showarticletitle{Asleep at the keyboard? assessing the security of
  github copilot’s code contributions}. In \bibinfo{booktitle}{\emph{2022
  IEEE Symposium on Security and Privacy (SP)}}. IEEE,
  \bibinfo{pages}{754--768}.
\newblock


\bibitem[Robinson et~al\mbox{.}(2022)]%
        {robinson2022}
\bibfield{author}{\bibinfo{person}{Joshua Robinson},
  \bibinfo{person}{Christopher~Michael Rytting}, {and} \bibinfo{person}{David
  Wingate}.} \bibinfo{year}{2022}\natexlab{}.
\newblock \bibinfo{title}{Leveraging Large Language Models for Multiple Choice
  Question Answering}.
\newblock
\newblock
\urldef\tempurl%
\url{https://doi.org/10.48550/ARXIV.2210.12353}
\showDOI{\tempurl}


\bibitem[Savelka et~al\mbox{.}(2023)]%
        {savelka2023}
\bibfield{author}{\bibinfo{person}{Jaromir Savelka}, \bibinfo{person}{Arav
  Agarwal}, \bibinfo{person}{Christopher Bogart}, {and} \bibinfo{person}{Majd
  Sakr}.} \bibinfo{year}{2023}\natexlab{}.
\newblock \showarticletitle{Large Language Models (GPT) Struggle to Answer
  Multiple-Choice Questions about Code}. In \bibinfo{booktitle}{\emph{15th
  International Conference on Computer Supported Education}}.
\newblock


\bibitem[Siddiq et~al\mbox{.}(2022)]%
        {10006873}
\bibfield{author}{\bibinfo{person}{Mohammed~Latif Siddiq},
  \bibinfo{person}{Shafayat~H. Majumder}, \bibinfo{person}{Maisha~R. Mim},
  \bibinfo{person}{Sourov Jajodia}, {and} \bibinfo{person}{Joanna C.~S.
  Santos}.} \bibinfo{year}{2022}\natexlab{}.
\newblock \showarticletitle{An Empirical Study of Code Smells in
  Transformer-based Code Generation Techniques}. In
  \bibinfo{booktitle}{\emph{2022 IEEE 22nd International Working Conference on
  Source Code Analysis and Manipulation (SCAM)}}. \bibinfo{pages}{71--82}.
\newblock
\urldef\tempurl%
\url{https://doi.org/10.1109/SCAM55253.2022.00014}
\showDOI{\tempurl}


\bibitem[Vaithilingam et~al\mbox{.}(2022)]%
        {10.1145/3491101.3519665}
\bibfield{author}{\bibinfo{person}{Priyan Vaithilingam},
  \bibinfo{person}{Tianyi Zhang}, {and} \bibinfo{person}{Elena~L. Glassman}.}
  \bibinfo{year}{2022}\natexlab{}.
\newblock \showarticletitle{Expectation vs. Experience: Evaluating the
  Usability of Code Generation Tools Powered by Large Language Models}. In
  \bibinfo{booktitle}{\emph{Extended Abstracts of the 2022 CHI Conference on
  Human Factors in Computing Systems}} (New Orleans, LA, USA)
  \emph{(\bibinfo{series}{CHI EA '22})}. \bibinfo{publisher}{Association for
  Computing Machinery}, \bibinfo{address}{New York, NY, USA}, Article
  \bibinfo{articleno}{332}, \bibinfo{numpages}{7}~pages.
\newblock
\showISBNx{9781450391566}
\urldef\tempurl%
\url{https://doi.org/10.1145/3491101.3519665}
\showDOI{\tempurl}


\bibitem[Vaswani et~al\mbox{.}(2017)]%
        {vaswani2017attention}
\bibfield{author}{\bibinfo{person}{Ashish Vaswani}, \bibinfo{person}{Noam
  Shazeer}, \bibinfo{person}{Niki Parmar}, \bibinfo{person}{Jakob Uszkoreit},
  \bibinfo{person}{Llion Jones}, \bibinfo{person}{Aidan~N Gomez},
  \bibinfo{person}{{\L}ukasz Kaiser}, {and} \bibinfo{person}{Illia
  Polosukhin}.} \bibinfo{year}{2017}\natexlab{}.
\newblock \showarticletitle{Attention is all you need}.
\newblock \bibinfo{journal}{\emph{Advances in neural information processing
  systems}}  \bibinfo{volume}{30} (\bibinfo{year}{2017}).
\newblock


\bibitem[Wermelinger(2023)]%
        {wermelinger2023using}
\bibfield{author}{\bibinfo{person}{Michel Wermelinger}.}
  \bibinfo{year}{2023}\natexlab{}.
\newblock \showarticletitle{Using GitHub Copilot to Solve Simple Programming
  Problems}.
\newblock  (\bibinfo{year}{2023}).
\newblock


\bibitem[Zong and Krishnamachari(2022)]%
        {zong2022solving}
\bibfield{author}{\bibinfo{person}{Mingyu Zong} {and} \bibinfo{person}{Bhaskar
  Krishnamachari}.} \bibinfo{year}{2022}\natexlab{}.
\newblock \showarticletitle{Solving math word problems concerning systems of
  equations with gpt-3}. In \bibinfo{booktitle}{\emph{Proceedings of the
  Thirteenth AAAI Symposium on Educational Advances in Artificial
  Intelligence}}.
\newblock


\end{thebibliography}

%%
%% If your work has an appendix, this is the place to put it.
% \appendix

% \section{Research Methods}

% \subsection{Part One}

% Lorem ipsum dolor sit amet, consectetur adipiscing elit. Morbi
% malesuada, quam in pulvinar varius, metus nunc fermentum urna, id
% sollicitudin purus odio sit amet enim. Aliquam ullamcorper eu ipsum
% vel mollis. Curabitur quis dictum nisl. Phasellus vel semper risus, et
% lacinia dolor. Integer ultricies commodo sem nec semper.

% \subsection{Part Two}

% Etiam commodo feugiat nisl pulvinar pellentesque. Etiam auctor sodales
% ligula, non varius nibh pulvinar semper. Suspendisse nec lectus non
% ipsum convallis congue hendrerit vitae sapien. Donec at laoreet
% eros. Vivamus non purus placerat, scelerisque diam eu, cursus
% ante. Etiam aliquam tortor auctor efficitur mattis.

% \section{Online Resources}

% Nam id fermentum dui. Suspendisse sagittis tortor a nulla mollis, in
% pulvinar ex pretium. Sed interdum orci quis metus euismod, et sagittis
% enim maximus. Vestibulum gravida massa ut felis suscipit
% congue. Quisque mattis elit a risus ultrices commodo venenatis eget
% dui. Etiam sagittis eleifend elementum.

% Nam interdum magna at lectus dignissim, ac dignissim lorem
% rhoncus. Maecenas eu arcu ac neque placerat aliquam. Nunc pulvinar
% massa et mattis lacinia.

\end{document}